\documentclass[letterpaper, 10 pt, conference]{ieeeconf}
\IEEEoverridecommandlockouts
\usepackage{xcolor}

\usepackage{cite}
\usepackage[pdftex]{graphicx}
\usepackage{amsmath}
\usepackage{url}
\title{\LARGE\bf Optical Navigation in Unstructured Dynamic Railroad Environments}

\author{Darius Burschka$^{1}$, Christian Robl$^{2}$ and Sebastian
Ohrendorf-Weiss$^{3}$%
\thanks{$^{1}$ Department of Computer Science, Technical University
Munich,  Germany, {\tt\small burschka@tum.de}}%
\thanks{$^{2}$M2C ExpertControl GmbH, Offenberg, Germany,\newline {\tt\small
Christian.Robl@m2cec.com}}%
\thanks{$^{3}$SBB AG, Bern, Switzerland,\newline {\tt\small
Sebastian.Ohrendorf-Weiss@sbb.ch}}}

\begin{document}

%

\maketitle
\thispagestyle{empty}
\pagestyle{empty}

\begin{abstract}

We present an approach for optical navigation  in unstructured, dynamic
railroad environments. We propose a way how to cope with the estimation of the
train motion from sole observations of the planar track bed. The occasional
significant occlusions during the operation of the train limit the available
observation to this difficult to track, repetitive area. This
approach is a step towards replacement of the expensive train management
infrastructure with local intelligence on the train for SmartRail 4.0.

We derive our approach for robust estimation of translation and 
rotation in this difficult environments and provide experimental
validation of the approach on real rail scenarios.

\end{abstract}

\section{Motivation} 

The increasing demand on public transportation requires an increase of
the train density, which begins reaching the capacity of the
conventional train infrastructure. The infrastructure based on balises
and signals has a fixed segment size that can accommodate just one train
with an empty segment in-between the trains. The increasing density
requires to switch to more flexible infrastructure, which is able to
localize the train within the train route and check the train
consistency. It is important that the entire train is leaving a specific
area and no train cars are left behind.  To exploit the potential of new
technologies, SBB, BLS, Schweizerische Südostbahn AG (SOB), Rhätische
Bahn (RhB),  Transports publics fribourgeois (TPF) and the Association
of Public Transport (VöV) have joined forces in the SmartRail 4.0
program.  With the SmartRail 4.0 program, the Swiss Railways want to
further increase capacity and safety, use the railway infrastructure
more efficiently, save costs, and maintain the competitiveness of
the railways in the long term.  SmartRail 4.0 has the ambition to
achieve a substantial improvement in the core of railway production.
Railway production includes all resources, systems and processes for
planning and safely executing movements on the railway infrastructure.
More capacity is to be made available on the existing track
infrastructure, for which a more precise and safe localization of
rail-bound vehicles is absolutely necessary.

Localization is essential in the field of control and safety technology
for the railway operation. Today, the localization of rail-bound
vehicles is based on the artificial infrastructure consisting of track
clearance sensors, balises in the track or signals in the event of a
fault.  Disadvantages are the high costs of these outdoor facilities,
suboptimal use of the line capacity due to the necessity of segment-wise
operation.  Today, absolute localization is only solved for specific
use-cases in certain areas, for example in the ETCS Level 2 corridors at
a speed above 40 km/h.
\begin{figure}[t!]
\centering
\includegraphics[width=0.48\columnwidth]{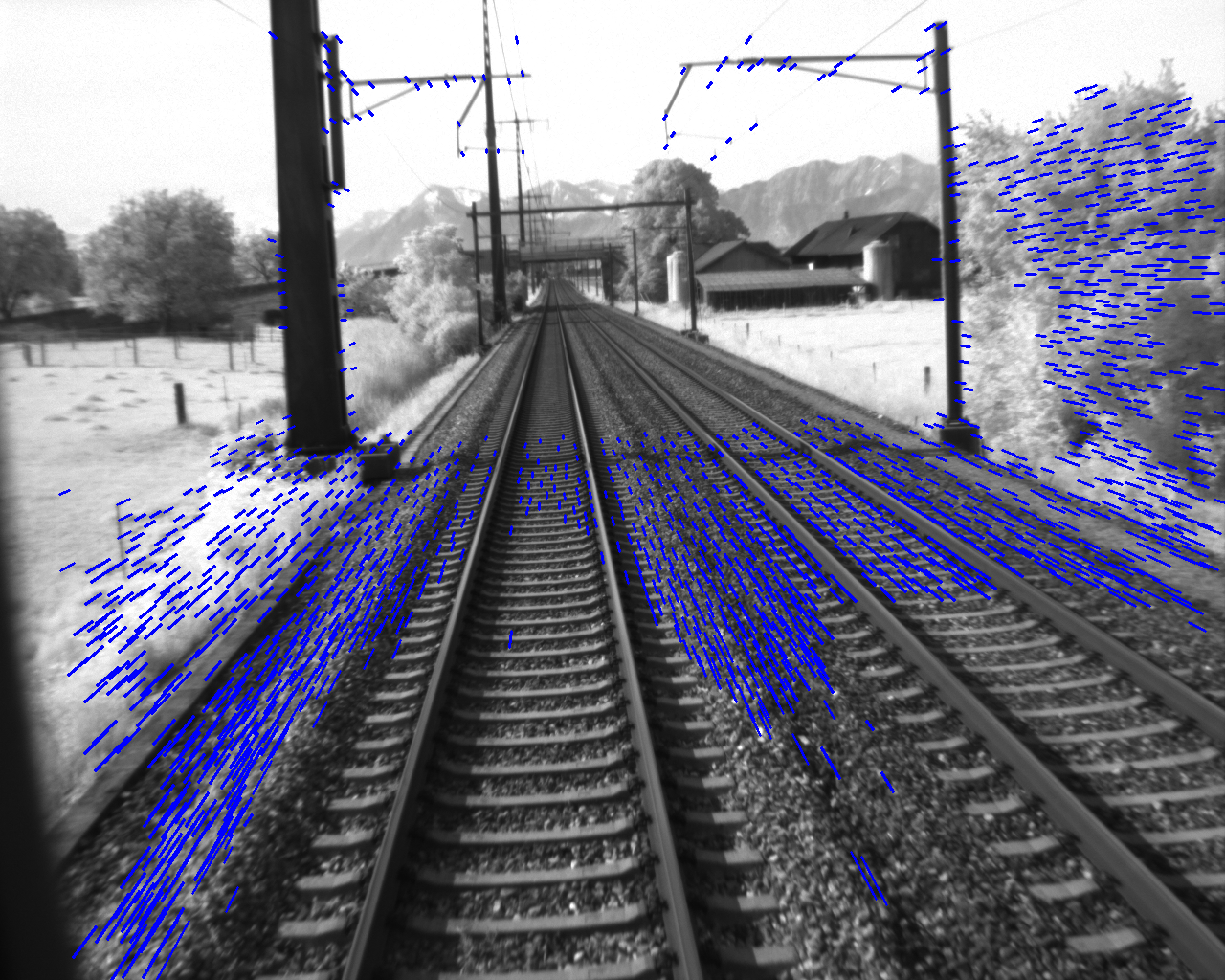}
\includegraphics[width=0.48\columnwidth]{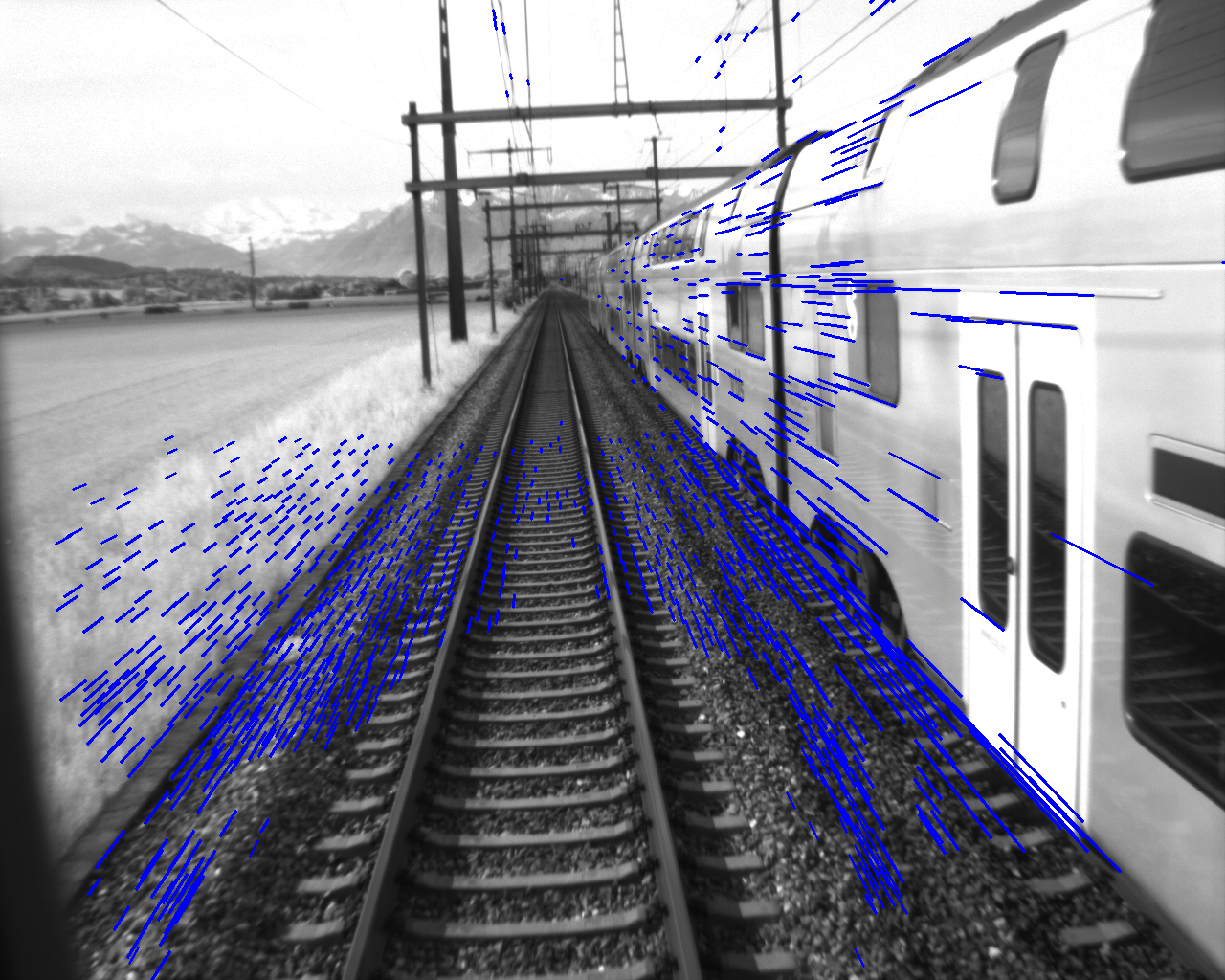}
\caption{\label{teaser::fig} Strong dynamic occlusions in railroad
scenarios:  (left) tracking data in static 
environment; (right) navigation under strong dynamic occlusion}
\end{figure}

In order to be able to remove the additional track-side infrastructure,
the precise and safe localization unit must be available on the vehicle.
The challenge that requires additional research beyond the current state
of the art method is that such a unit needs to provide verifiable
results, which does not allow an application of Deep Learning methods
and forbids even an application of standard methods, like RANSAC, due
to its randomized processing with slightly varying results on same
inputs. Additionally, the reliable static background information is
limited only on the planar surfaces with highly self-similar, planar and
repetitive structure of the gravel. This prevents applications of
stereo-based system, which rely on clear 3D boundaries of planes, and of
traditional sparse systems like ORBSlam, because the local features are
not unique. It also requires powerful Future Railway Mobile Communication System
(FRMCS) technology to monitor train integrity and send the exact
position to the central interlocking.~\cite{germann}.  The development
of localization is a decisive factor in the evolution of digitalization
in the field of control and safety technology of railway systems.
Additionally, the use of localization can also trigger a performance
boost in today's digital interlocking technology. With the integration
of the new, precise localization technology, the operational performance
of today's control and safety technology can be increased in multiple
domains. A standstill detection would make it possible to avoid track
closures due to a lack of slip paths and thus achieve a more efficient
use of the existing facilities.  If today's infrastructure- and
odometry-based localization can be further developed into a continuous,
object-side, autonomous SIL4 localization, enormous opportunities will
open up for increasing efficiency and safety in a large number of
railway applications. 

We aim to make it possible to operate within the absolute braking
distance in so called “Moving Blocks”.  Consequently a more efficient
handling of rail traffic is possible and leads to more track capacity on
the very dense rail network used. 

There are three main obstacles for the safe and precise localization of
rail-bound vehicles:
\begin{enumerate}
\item Finding a sensor-combination and -fusion for a highly available, secure, safe and precise localization.
\item Obtain a SIL4 approval for the new localization system through the relevant certification bodies.
\item Secure interoperability within the European railways and setting international standardization.
\end{enumerate}

The approach for localization described in this paper is only one out of several possible approaches that are taken into account in SmartRail 4.0.


\subsection{Rail-specific Navigation Problems}

The applications for trains and rail infrastructure have
to comply with highly-restrictive standards, like the CENELEC standards
(EN 5012x, EN 50657), in order to be certified by the authorities.
RAMS (Reliability, Availability, Maintainability, Safety) requirements
have to be fulfilled  to reach the required SIL (system integrity level).
Therefore, safety critical applications such as a SIL 4 localization of
track bound vehicles must be redundant (to reach availability) and need
to meet diverse constraints in using deterministic algorithms (to reach
safety level). Thus a use of machine learning or artificial intelligence
approaches is not suitable for such systems.  As we mentioned already
earlier, the requirement to provide the same accurate measurement
together with an information about the achieved accuracy from a specific
image set, does not allow to use any probabilistic methods. It prohibits
even the use of common techniques, like RANSAC, for model verification.

\subsection{Related Work}
Optical navigation systems can be categorized based on the input data that
they rely on. There exist many commercial systems~\cite{realsense,zed}
that provide optical navigation data from 3D reconstructions in a
binocular camera or camera-projector system. These approaches require
static 3D structures like trees, houses or other objects in the scene
that can be tracked over time. There are active 3D navigation systems
mostly for door navigation, like RealSense camera, and outdoor binocular
stereo systems, like ZED. As we can
see in Fig.~\ref{teaser::fig} such systems fail occasionally in the
specific application field of a railroad scenario, because the strong
occlusions by other trains passing on parallel tracks limit the
available data to just the track bed in front of the train. The other
type of navigation systems rely on the image information itself and can
be subdivided in dense systems using the information of every pixel in the
image~\cite{lsdslam} or matching significant points representing strong
multi-directional brightness changes in the image~\cite{Zisserman,
zinf}. We do not consider learning approaches in our framework, because
the resulting navigation system needs to undergo a strict verification
process to be applied on trains and the current learning approaches do
not meet this requirement. 

There exist many optical navigation frameworks developed for the field of service
robotics~\cite{davison,burschka,lsdslam} and for outdoor navigation~\cite{newmann,
burschka, orbslam}. These systems have in common that they rely on matching
of local image information over a time-sequence of images that creates
the so-called {\em optical flow}, which is analyzed for its rotational
and translational effects~\cite{Zisserman}. These approaches fail in
many situations in railroad environments, because of the strong
self-similarity of structures in the track bed, which is often the only
reliable reference to the static environment. This required us to
develop a different matching system that copes with this unstructured,
repetitive property of the environment. It is presented in
Section~\ref{mouse:sec}.

Current monocular or stereo systems when applied on trains also suffer from
the unstructured, repetitive environment, drifting gains and from aliasing to due the
limited frame rate (20fps) at a max speed of 52,4 km/h~\cite{siegwart}.
In addition scenarios with other dynamic objects like cars and other trams or fast switching between shadow and sunlight limits the system performance in ~\cite{siegwart}.  
Internal projects at SBB used cameras of maintenance vehicles to
identify landmarks on or close to the track (e.g. balises, signs).
However they are using machine learning algorithms which cannot be
applied on safety critical applications like train localization. Due to
the high RAMS requirements most of such vision based systems are applied
as assistance systems only, today, even if autonomous driving would be
the final goal~\cite{siemens}. The tram driver still has to override
such assistance systems in order to avoid unnecessary emergency braking.
Applying navigation assistance systems from the automotive domain fails
due to the different environment and safety cases in rail. Odometry supported by wheel encoder like those used in automotive suffer from the high slip of rail vehicles (modern locomotives drive intentionally with slip.)    

Many of the available systems are not able to provide any additional
information about the quality of the currently estimated pose change of
the camera. If the result is supposed to be fused in a fusion framework
like a Kalman-Filter then the resulting covariance needs to be kept at a
constant, worst-case level. We propose to extend the navigation
approaches to provide the current uncertainty in the estimation of the
pose in parallel to the navigation information. The accuracy may
strongly vary based on the distance to the observed objects and their
distribution in the camera image. The extension based on our
work in~\cite{elmarIros} allows to estimate the quality of the
processing (QoS) for each navigation step enabling a better convergence
of the fusion framework. The proposed processing is described in
Section~\ref{nav:sec} in more detail.

We present our approach, how to estimate the motion properties in
Section~\ref{nav:sec}. Section~\ref{mouse:sec} presents our approach,
how to estimate robustly the metric motion of the camera in the highly
unstructured area of the track bed. The possible drifts of the presented
system are compensated through occasional information from global
infrastructure, which is presented in Section~\ref{nav:sec}.We present in
Section~\ref{result:sec} the achieved  accuracy in the motion estimation
and metric measurements. We conclude with some final evaluation of the
achieved system properties and discuss our further directions in
Section~\ref{conclusion:sec}.

\section{Approach}


Visual localization provides information about motion of the camera
relative to structures in the surrounding environment through direct
observation of the changes in their projected position in the images.
This prevents accumulation of position and orientation errors as long as
the same global features can be kept visible. The position is extracted
through processing of the image information and the pose (position and
orientation) change is calculated with frame-rates in the range of
30-120Hz. This frequency limits the maximal possible dynamic motion of
the system to changes occurring with a frequency smaller than half of
that frame-rate.  

In contrast, sensors like inertial units (IMUs) rely solely on physical
effects within the sensor as a response to applied velocities and
accelerations. They provide a significantly higher measurement rate,
which can reach for an inertial unit values around 800-1000Hz. Small
errors in the estimate due to noise or external disturbance  cannot be
compensated here through a global reference. They cause drifts that are
accumulated during the integration of the consecutive measurements. A
visual system exposes similar errors but with a significantly slower
frequency, when a reference used to measure the current position needs
to be changed to a new landmark (hand-off
problem)~\cite{darius_loc_chapter}. 

We can see in Fig.~\ref{teaser::fig} that the reliable area of the
tracks that can be used for navigation does not provide unique matches
in the tracking system that provides the information about  the motion
of the train. Local matching strategies that work reliably for flying
systems and in the automotive domain cannot be applied in railroad
environments due to this strong self-similarity of the local features.

\begin{figure}[ht]
\centering
\includegraphics[width=9cm]{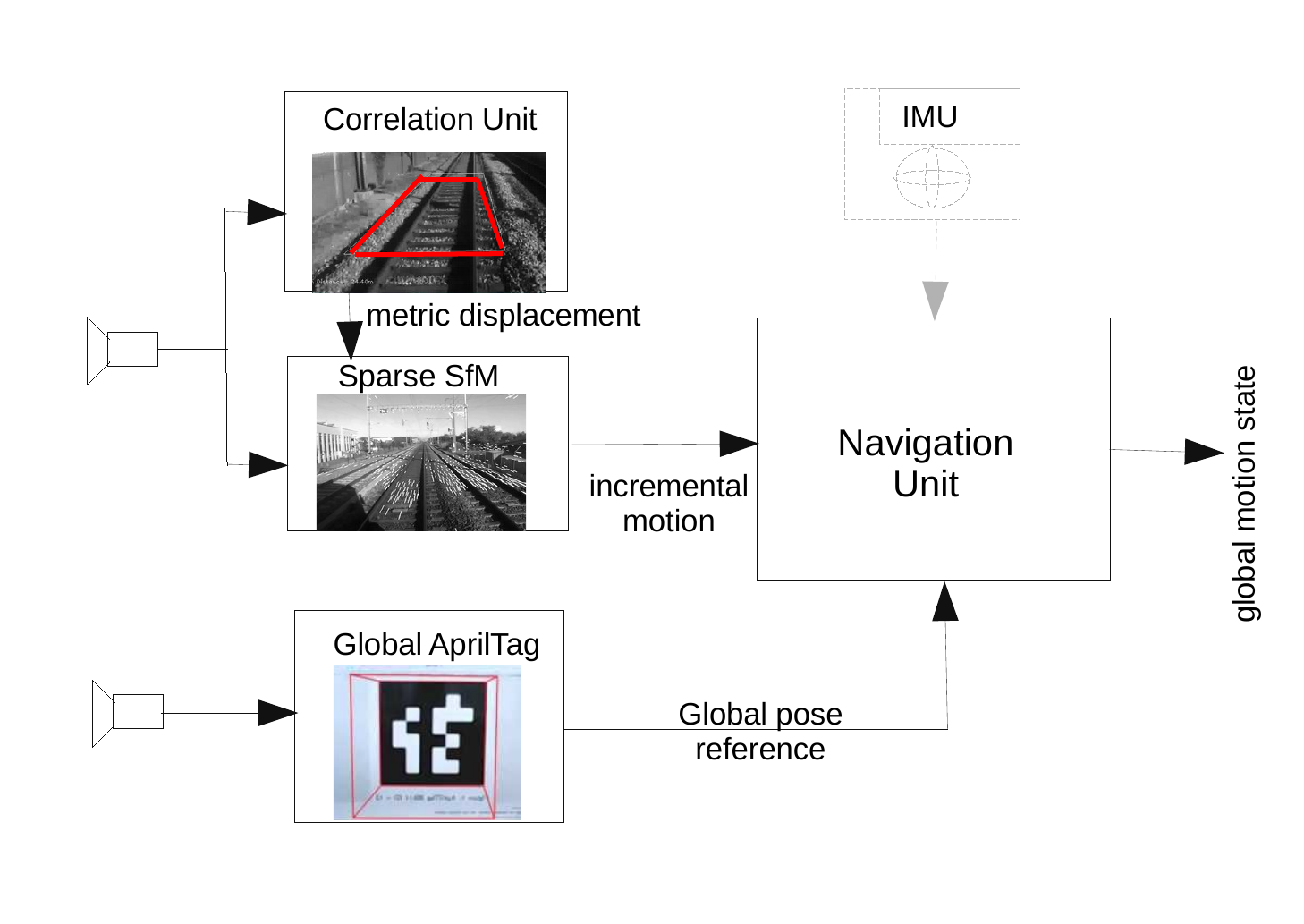}
\caption{\label{system:fig} System architecture of the planned goal
navigation system.(SfM: Structure from Motion)}
\end{figure}

We propose a final system architecture depicted in Fig.~\ref{system:fig} to
solve this problem. The navigation unit fuses the information from a
point-based structure-from-motion (SfM unit - Section~\ref{nav:sec})
with a unit correlating large areas of the tracks to estimate robustly
the metric motion of the train (correlation unit). The dynamic motion
state of the train is currently estimated only from fusion of the
optical unit with the Kalman Filter prediction. We plan to extend it
with the information provided by an additional inertial unit (IMU) to
allow capture of higher dynamic motions of lighter train setups.

Our system is calculating the pose changes from a monocular image
sequence. This sequence is passed to a correlation unit that estimates
the metric translational motion of the train from the motion of an image
template in the track area between the images of the sequence. The
details of the processing are presented in Section~\ref{mouse:sec} in
more detail. The rotational parameters and the direction of motion is
calculated from a modified SfM module, where additionally the accuracy
of the current navigation result is estimated. It is important for
correct fusion in the Fusion Unit and for the planned certification of
the system. This processing is presented in Section~\ref{nav:sec}.

The presented system cannot avoid long term drifts, because the
correlation unit and the SfM module rely only on local features that can
be used as reference only in limited space.  Our system uses an
additional long focal length camera that identifies
April-Tags~\cite{aprilurl} placed instead of the usual identifiers along
the track. These tags are used to compensate possible drifts in the
navigation unit. They provide geo-tagged information about the position
of the train in the world.

The navigation unit~(Fig.~\ref{system:fig}) can further optimize the
calculation of the distance by freezing the reference frame~$\cal{I'_t}$
(key-frame)
for a number of following frames, if the estimated velocity is slow. Since
the distance travel is the integral of the responses from the optical
correlation, small detection errors usually integrate to increasing
drifts in the distance. Switching to the key-frame-processing results in
the detection errors appearing as noise overlayed over the true distance
instead of appearing as accumulated drift (Section~\ref{mouse_result}).

\subsection{Robust Estimation of Metric Motion Parameters}
\label{mouse:sec}

Conventional Visual SLAM approaches use the information from a sparse
point matching system in the camera images. The points are {\em tracked}
between the image pairs from the sequence or {\em matched} based on the
local information in the neighborhood of the points. The difference is
that while {\em tracking} assumes a local search around the expected
position, in which a local image patch is searched, {\em matching}
allows larger changes in the image position, because each point is
described by a more or less complex description (SIFT~\cite{lowe},
AGAST~\cite{agast}).

\begin{figure}[ht]
\centering
\vspace{2ex}
\includegraphics[width=0.8\columnwidth]{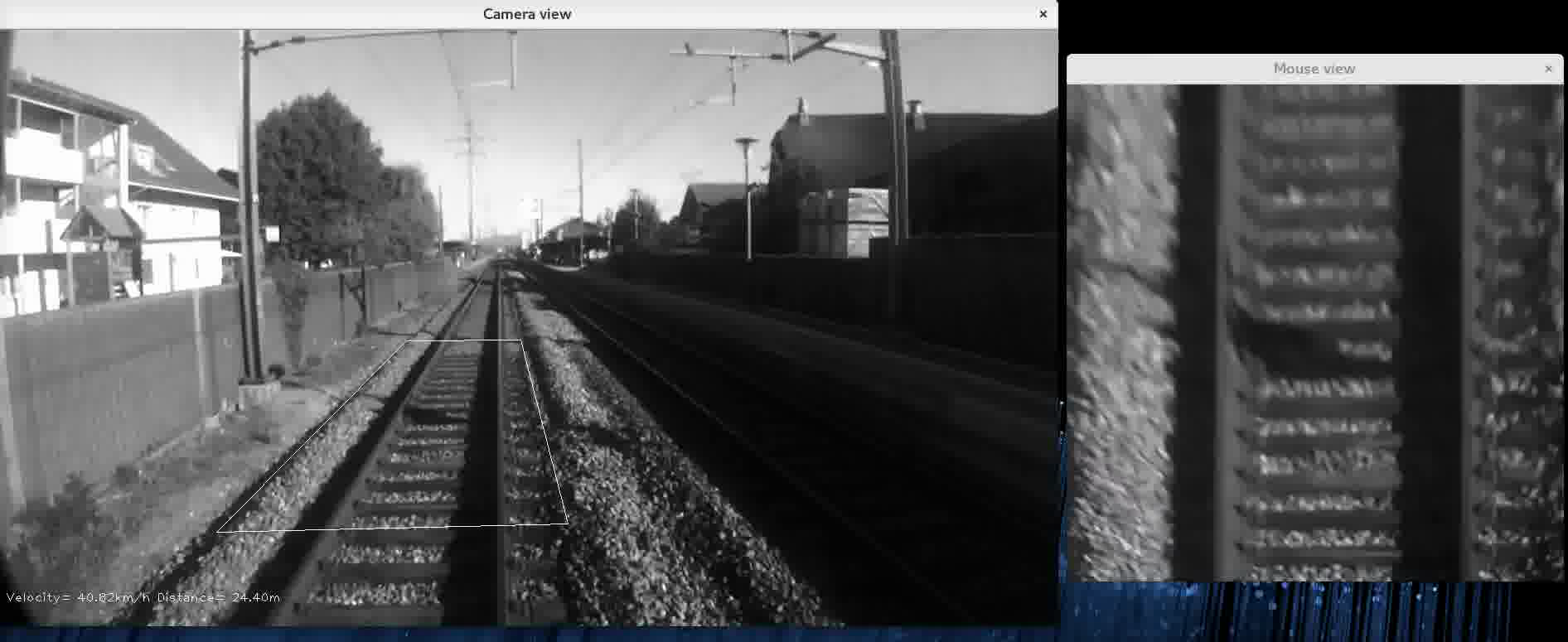}
\caption{\label{mousereg:fig} The rectangular region shown in the left
image is rectified to the ``top-view'' image shown on the right. A
template in this image is searched in the consecutive image rectified in
the same way.}
\end{figure}

While this processing works in most
flying and automotive environments, we need to be able to match the
information in the area of the tracks with a very strong self similarity
that leads to many mismatches between the frames. We increase the
uniqueness of the local environment by growing the local region to a
large area shown in the Fig.~\ref{mousereg:fig}. We try to match this
template in the consecutive image using a Sum-of-Square-Differences
(SSD) method from OpenCV. We refer to this module because of the
similarity to an optical computer mouse as ``Train Mouse''.

A homography matrix~$\tilde{H}$ that is used to calculate the rectified
image~$\cal{I'}$ in Fig.~\ref{mousereg:fig} right has the generic
structure~(\ref{hom:eq}):
\begin{equation}
\label{hom:eq}
\tilde{H}=\left(\tilde{R}+\frac{\vec{T}\vec{n}^T}{d}\right)
\quad \rightarrow \quad\cal{I'}=\tilde{H}\cdot \cal{I}
\end{equation}
The rotation matrix~$\tilde{R}$ describes the rotation between the
current orientation of the physical camera and the top-view orientation
of the rectified view. The vector~$\vec{T}$ describes the translation
between the images, which is zero in our case. Therefore, plane normal
vector~$\vec{n}$ of the tracks and the distance of the camera to the
tracks~$d$ become irrelevant here. We use the homography to rotate the
camera image~$\cal{I}$ to the top-view~$\cal{I'}$ orientation.

We search for a rectangular template with the size~$(x',y')$ from
the~${\cal{I'}}_t$  region of the first image in the
corresponding region~${\cal{I'}}_{t+1}$ using the SSD~template matching
method that searches for the maximum of the
function~(\ref{pixelssd:eq}):
\begin{equation}
\label{pixelssd:eq}
f(x_p,y_p)=\sum_{x',y'}({\cal{I'}}_t(x',y')-{\cal{I'}}_{t+1}(x_p+x',y_p+y'))^2
\end{equation}

The estimated displacement~$(x_p,y_p)_t$ from the maximum response
of~$f(x_p,y_p)$ estimates the horizontal and vertical
image motion of the template between the images. This measures a pixel
accurate shift of the template between the images. The search for the
correct displacement for the current~$(x_p,y_p)_t$ can be accelerated by
using a prediction of these values. In a generic case, the system needs
to check the entire possible range of~$\{x_p,y_p\}$ that covers the
entire possible velocity profile. This is a computationally intensive
operation. Due to the high inertia of the
train, these value change only little between consecutive frames.
We can reduce the search for the correct placement of the template only
to a small band around the previous~$(x_p,y_p)_{t-1}$ values.

We can calculate a more accurate displacement of the template between
the images by applying a sub-pixel alignment of the templates. If the
remaining change between both images is under 1~$\left[\mbox{pixel}\right]$ then
we can use the Taylor series expansion to explain the brightness change
at a specific pixel~${\cal{I'}}(x,y)$ to:

\begin{gather}
 {\cal{I'}}_t(x+\delta x.y+\delta y)\approx\\ \nonumber
 {\cal{I'}}_t(x,y)+
 \frac{\partial {\cal{I'}}_t(x,y)}{\partial x}\delta x+
 \frac{\partial {\cal{I'}}_t(x,y)}{\partial y}\delta x
\end{gather}

If we assume that the new image~${\cal{I'}}_{t+1}$ is a result of a
sub-pixel motion~$(\delta x,\delta y)$ then we can estimate from the
equation:
\begin{gather}
 \nonumber{\cal{I'}}_{t+1}(x.y)-{\cal{I'}}_t(x,y)\approx\\ 
 \label{partial:eq}
 \frac{\partial {\cal{I'}}_t(x,y)}{\partial x}\delta x+
 \frac{\partial {\cal{I'}}_t(x,y)}{\partial y}\delta
 x=\vec{\cal{G}}^T\cdot\delta \vec p=
 ||\vec{\cal{G}}||\cdot||\delta\vec p||\\
 \mbox{with} \quad\vec{\cal{G}}=\left(\frac{\partial
 {\cal{I'}}_t(x,y)}{\partial x}, \frac{\partial
 {\cal{I'}}_t(x,y)}{\partial y}\right)^T\nonumber
\end{gather}
We see that once we calculated the gradient vector~${\cal{G}}$ from
the previous image, we can calculate the sub-pixel update of the motion
in horizontal and vertical direction~$(\delta x.\delta y)$ by
decomposing the motion~$||\vec{\delta p}||$ along the gradient according to the
horizontal and vertical ratios of~$\vec{\cal{G}}$. 

We calculate the resulting shift as an average of responses within the
template. It is obvious from~(\ref{partial:eq}) that only pixels with
a difference in brightness between the images contribute to the motion
estimation. We reduce the sensitivity to noise by using only pixels with
the gradient above a threshold
$||\vec{\cal{G}}||>\epsilon_G$, which is tuned depending on the expected
camera noise.

The resulting average image motion~$(\Delta x,\Delta y)$  can be linearly
scaled to the forward and side-wards metric velocity with knowledge about
the mounting height~L above the ground.
The metric values of the forward velocity~$v_l$ and the side-wards
motion~$v_s$
(due to curves in the route) can be computed from {\em similar triangles}
relation between the camera projection on the image plane and the
relation of the height~L of a rectified camera providing the
image~$\cal{I'}$ to:
\begin{gather}
\Delta x_i=x_p+\delta x, \quad \Delta y_i=y_p+\delta y\nonumber\\
\label{img_mot:eq}
v_l=\frac{L\cdot p_y}{f\cdot t_f}\Delta y_i, \quad
v_s=\frac{L\cdot p_x}{f\cdot
t_f}\Delta x_i
\end{gather}

Possible changes in the orientation of the camera  image~$\cal{I'}$
scale it with the focal length~$f$, the metric pixel-size~$(p_x,p_y)$
and the time interval between two frames~$t_f$ as it is shown
in~(\ref{img_mot:eq}). The improvement achieved with the extension to
the sub-pixel accuracy is shown in Section~\ref{mouse_result}.

A possible error in the estimate of the traveled global
distance~$(x_g,y_g)$ can occur due to the noise in the
brightness information~${\cal{I'}}_t(x,y)+\nu_i$. Since the global shift
is an integral (sum) of the consecutive steps~$(\Delta x,\Delta y)$, the
error accumulates fast in each step. The resulting shift in
each step is estimated as an average response of all significant
brightness changes within the templates. The statistical distribution of
the error helps to reduce the error in the final estimates. This can be
pushed even further by tracking a template not only between consecutive
images but over a longer period of time. The reference template from the
original image, which we will refer to as {\em keyframe} in the
following text, is used to estimate the shift in multiple following
frames. It is done until the template moves out from the area warped in
the convolution step above. This processing introduces the brightness
noise-related error only once in the navigation process, instead of being
added multiple times with each new delta step. The length of the
sequence, in which a keyframe can be used, depends directly on the
current speed of the train. We will see in Section~\ref{nav:sec} that
this processing has an additional advantage on the motion estimation
process.

A significant advantage of adding a separate estimation of the forward
motion is the possibility of estimation of the typically unobservable
motion error~$\sigma_z$ along the optical axis~Z. We are able to estimate this
error from the~$\Delta y_i$ responses of all~N contributing pixels in the
template with the property~$||{\cal{G}}||>\epsilon_G$ to:
\begin{gather}
\label{sigma:eq}
\sigma_z^2=\frac{1}{N}\sum_{i\in ||\vec{\cal{G}}||>\epsilon_G} (\Delta
y_i-\overline{\Delta y})^2 ,\; \overline{\Delta y}=\frac{1}{N}\sum_{i\in ||\vec{\cal{G}}||>\epsilon_G} \Delta y_i
\end{gather}

\subsection{Robust Key-frame-based Monocular Motion Estimation}
\label{nav:sec}

\begin{figure}[ht]
\includegraphics[width=\columnwidth]{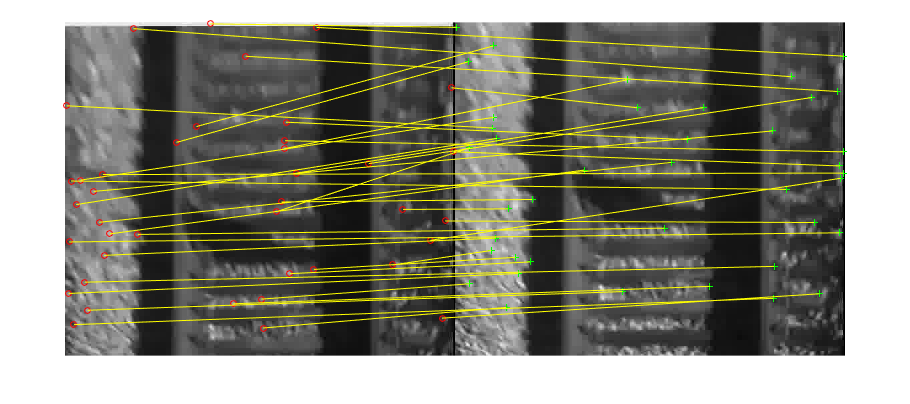}
\caption{\label{similarity:fig} Strongest matching candidates are often
not the correct correspondences for feature points in track area.}
\end{figure}

The {\em keyframe processing} introduced in the previous section
improves also the accuracy of the estimation of the direction of motion
in the monocular Essential matrix decomposition~\cite{Zisserman}. Our
problem with the matching of features between images of the sequence is
the significant self-similarity of the observed features
(Fig.~\ref{similarity:fig}). Typical matching algorithms like SURF,
BRISK, KAZE, find multiple matching candidates for a tracked point.

\begin{figure}[ht]
\centering
\includegraphics[width=0.7\columnwidth]{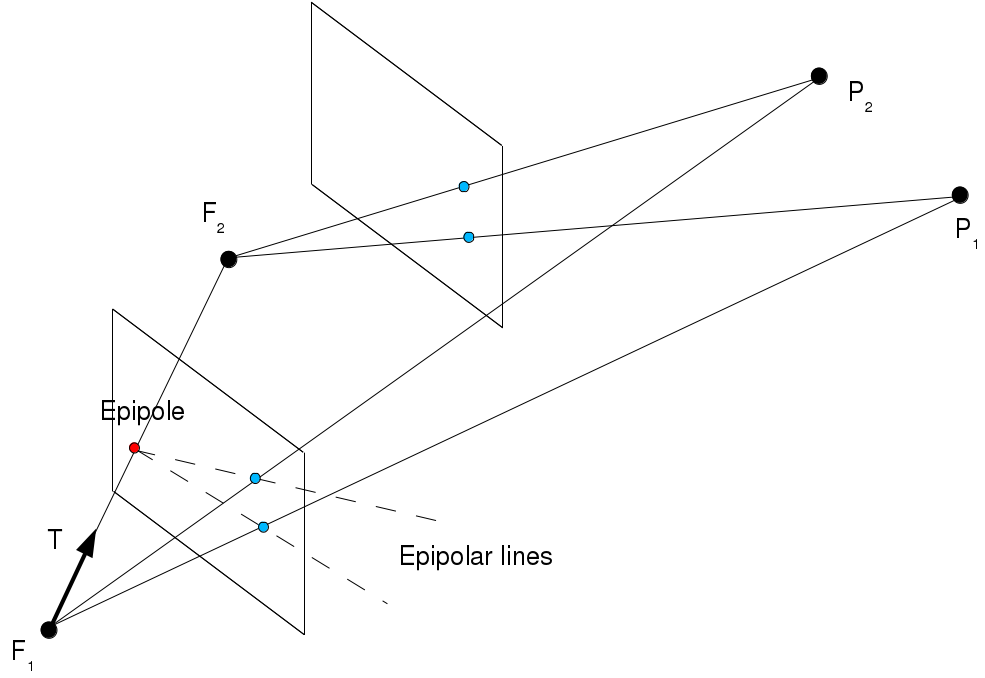}
\caption{\label{epipole:fig} While the camera is moving along the
vector~$\vec{T}$, the tracked points move along the dashed epipolar lines
in the second image frame.}
\end{figure}

The selection of the correct matching candidate can be largely
simplified. Since the previous optical correlation step found the planar
direction of motion, which represents the first~$T_x$ and the last~$T_z$
parameters of the motion vector, we can estimate the horizontal position
of the epipole in the image. The direction of the motion
vector~$\vec{T}$ defines the position of the intersection point of all
optical flow lines, which are segments of the corresponding epipolar
lines (see Fig.~\ref{epipole:fig}).

Our system uses the predicted value for the rotation matrix~R that is
calculated in the fusion framework of the {\em Navigation Unit}
(Fig.~\ref{system:fig}). We rotate all matched points~$\vec{p_i}=(u_i,v_i,f)^T$
by this matrix to a rotation compensated version~$\vec{p}'_i$:
\begin{equation}
  \vec{l}=\tilde{R}^T\cdot\vec{p_i}, \quad
  \underline{\vec{p}'_i=\frac{f}{l_z}\cdot \vec{l}}
\end{equation}
The resulting optical flow has just the translational component, which
intersects in the expected epipole. Since the rotation is just a
prediction, we allow the optical flow lines to deviate by a small pixel
value from this epipole. An example for the compensated optical flow
field can be found in Fig.~\ref{teaser::fig}. We choose the matches from
the matching pool, which point towards the expected epipole.

Once the correct matches between features in both images are found, we
estimate the new corrected~$\tilde{R}$ and the direction of motion
vector~$\vec{T}$ using processing similar to standard {\em calc\_pose()} method from
OpenCV without the RANSAC part. The filtering was done before in a
deterministic way.  The proposed novelty is the way, how we additionally filter the
wrong correspondences based on the expected epipole above. The solution
becomes ambiguous especially in strongly limited visible space without
this processing. 

An important final step in the processing is the calculation of the
covariance of the result. We estimated the metric~$\sigma_z$ component
already in~(\ref{sigma:eq}). We estimate the remaining two components by
estimating the distances, how far the lines associated with the flow segment 
miss the epipole. For an i-th flow vector with start and
end-point~$(\vec{p}_{si},\vec{p}_{ei})$, we can estimate the epipole
point~$\vec{x}_e$ to:
\begin{gather}
  \vec{k}_i=\left(\begin{array}{c} k_u\\
  k_v\end{array}\right)=\vec{p}_{ei}-\vec{p}_{si}\nonumber, \quad
  \vec{n}_i=\left(\begin{array}{c} -k_v\\ k_u\end{array}\right)\\
  \tilde{A}=\left(\begin{array}{c} \vec{n}_1^T\\.\\.\\
  \vec{n}_k^T\end{array}\right),\quad
  \vec{b}^T=\tilde{A}\cdot\left(\vec{p}_{s1},
  \dots,\vec{p}_{sk}\right),\quad
  \underline{\tilde{A}\cdot\vec{x}_e=\vec{b}}
  \label{pseudo:eq}
\end{gather}

The epipole position~$\vec{x}_e$ can be estimated using a pseudo-inverse
of the non-square matrix~$\tilde{A}$ in~(\ref{pseudo:eq}).
An essential information for a fusion in the {\em Navigation Unit} is the
covariance of the estimated value. It helps to assess the current
uncertainty in the measurement.
We calculate the
closest distance~$\vec{\delta p}_i$ for flow optical flow-line~$(\vec{p}_{si},\vec{p}_{ei})$
from~$\vec{x}_e$ using~$\Delta x$ result from~(\ref{img_mot:eq}) for a
scaling from pixel-values to meters as:
\begin{equation}
\vec{\delta_p}_i=\frac{\Delta x\cdot p_x}{f}\left[\vec{n}_i^T\cdot\left(\vec{x}_{si}-\vec{x}_e\right)\right]\cdot\vec{n}_i
\end{equation}

The resulting covariance matrix~$\tilde{P}$ in the xy-plane from~k optical flow lines is
constructed as:
\begin{equation}
\tilde{P}=\frac{1}{k}\sum_{i=1}^k \vec{\delta_p}_i\cdot \vec{\delta_p}_i^T
\end{equation}

The {\em keyframe processing} helps similar to the previous chapter to
reduce the error while switching to new references by significantly
reducing the number of switches. Additionally, the flow line-segments
in the images become longer. If we assume a constant detection error for
the flow endpoints in the images then longer lines are less sensitive to
orientation changes due to the detection error.

\subsection{Drift Compensation from Global Landmarks}
\label{drif:sec}

Since (visual) odometry or IMUs only provide a relative localization, global landmarks are required for getting world coordinates. These relative algorithms also suffer from an accumulative offset and an unknown initial condition of the real value that leads to a drift from the real position. To compensate for that drift other sensors needs to be included in the sensor system. Using GNSS is the most promising approach, however in areas without GNSS coverage (e.g. in tunnels or valleys) another approach could be helpful. As such an alternative vision based solutions such as April tags (or similar tags or signs) mounted on the poles of the catenary, whose position is known in world coordinates with an accuracy within 10 cm, can be successfully used. The global pose of the camera relative to the tags can be easily derived. Further vision based approaches could be using other global and fixed landmarks of the environment (e.g. points). 

\section{Experimental Results}
\label{result:sec}

\begin{figure}[ht]
\centering
\includegraphics[width=0.7\columnwidth]{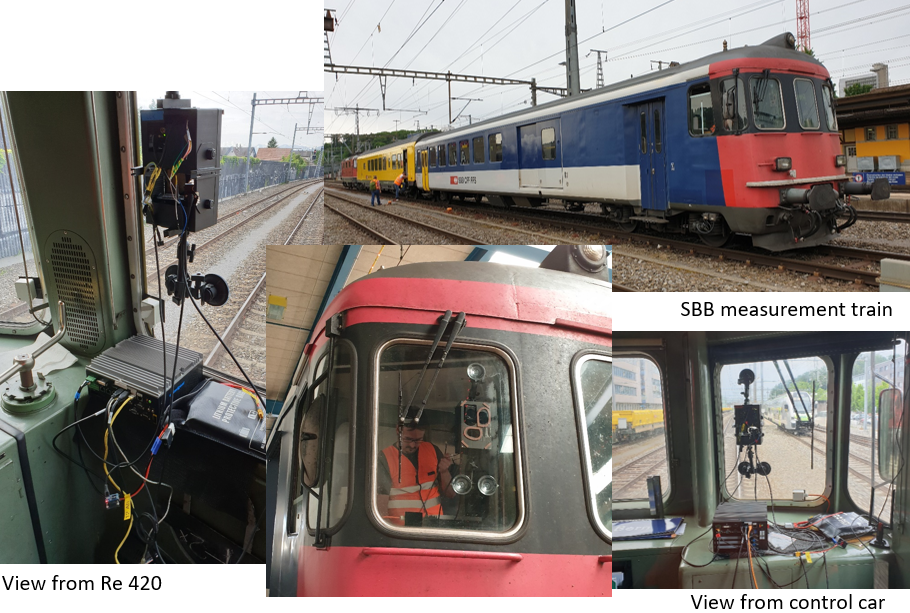}
\caption{\label{mtrain:fig} SBB measurement train and camera setup in locomotive (Re 420) and control car}
\end{figure}
The setup up for the trial was a specific train run organized by SBB
where cameras were installed on the windscreens of the locomotive and
the control car (see Fig.~\ref{mtrain:fig}). There were two cameras
each, one for the ``Train Mouse'' and one for the April tags mounted on
the poles for the catenary. The 10 bit NIR cameras with a resolution of
1280 by 1024 pixels were used at a frame rate of 60 fps. The cameras
were supported by IR-illuminators to overcome tunnels. All camera were
calibrated using the Bouquet toolbox with a 9x6 checkerboard with 8cm by
8cm tile size. The SBB telecom measurement wagon in the middle of the
train composition provided the position reference as it was equipped
with a DGNSS solution combined with an IMU (high performance ring-laser
gyro). The approx. 20 minutes trip between Ostermundigen and Thun were
repeated four times in order to reproduce results with different speeds
up to 140 km/h and different scenarios (e.g. occlusion by other trains).
This route was chosen because it also contains the 8km long fiber optic
sensing (FOS) test track and enabled a comparison of the results of the
different localization sensors. The route consists mostly of two
parallel tracks and 4 railway station were there are several points and
up to 6 parallel tracks.      

\subsection{Accuracy of the Correlation Approach (``Train Mouse'')}
\label{mouse_result}

\begin{figure}[ht]
\centering
\includegraphics[width=0.48\columnwidth]{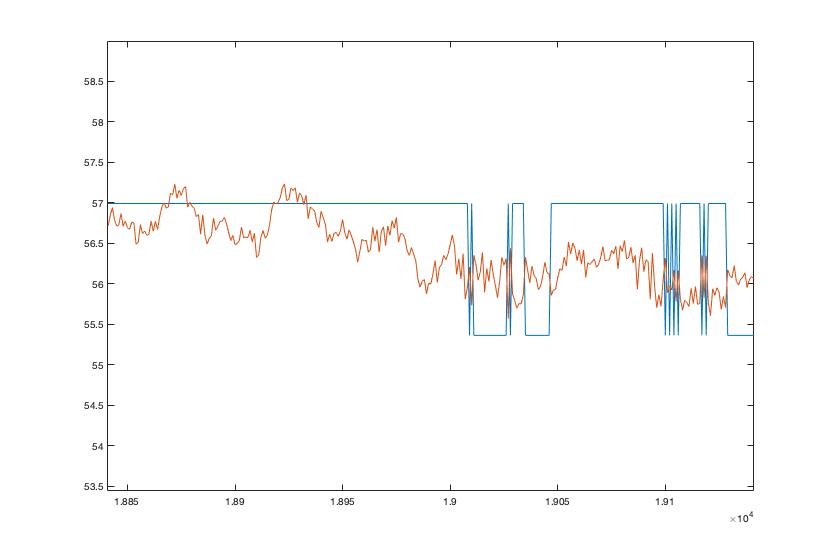}
\includegraphics[width=0.48\columnwidth]{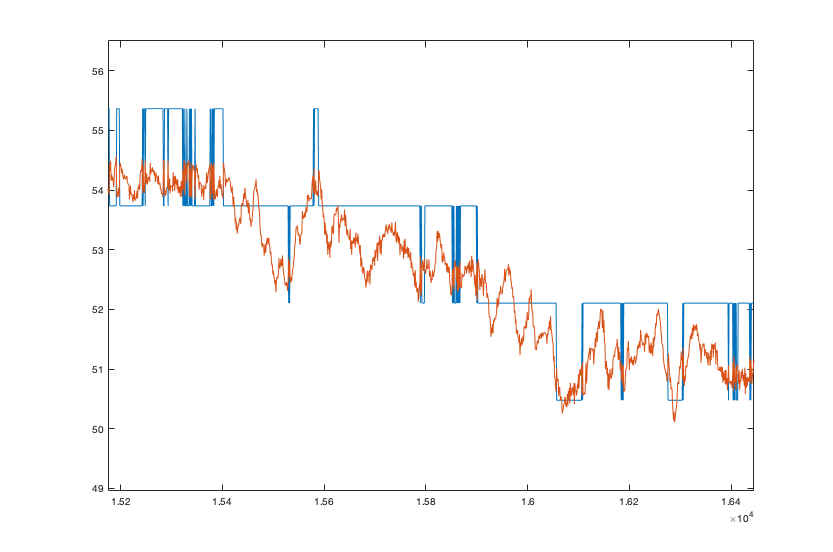}
\caption{\label{subpixel:fig}Comparison between velocity estimation
without (blue) and with (red) sub-pixel optimization.}
\end{figure}

Fig.~\ref{subpixel:fig} depicts the necessity to include the sub-pixel
motion estimation into the metric motion estimation system. This allows
an early notification about the train setting in motion even before the
human eye can observe it. It is also very important at higher velocities,
where a change of one pixel in motion between 2 frames corresponds to
multiple km/h at typical speeds of up to 140km/h.

The optical correlation system (``Train Mouse'') achieves an accuracy
beyond the capabilities of the existing mechanical and GNSS sensors
(Fig.~\ref{subpixel:fig}).  It is possible to see small velocity
changes, which can be used to analyze changes in the dynamic state of
the train, if multiple units are distributed over the length of the
train.  We can observe changes, e.g. due to oscillations of the train
control system, on the track. It is to our knowledge the first system in
the rail domain operating at velocities equal or higher than 140 kmh
because of successful solution of the matching problem through
correlation.

\begin{figure}[ht]
\centering
\includegraphics[height=2.5cm]{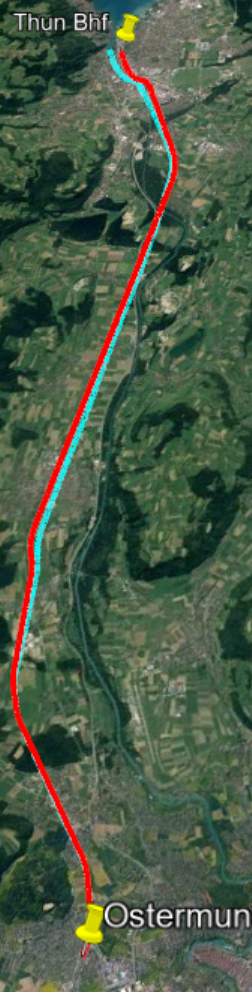}
\includegraphics[width=0.5\columnwidth]{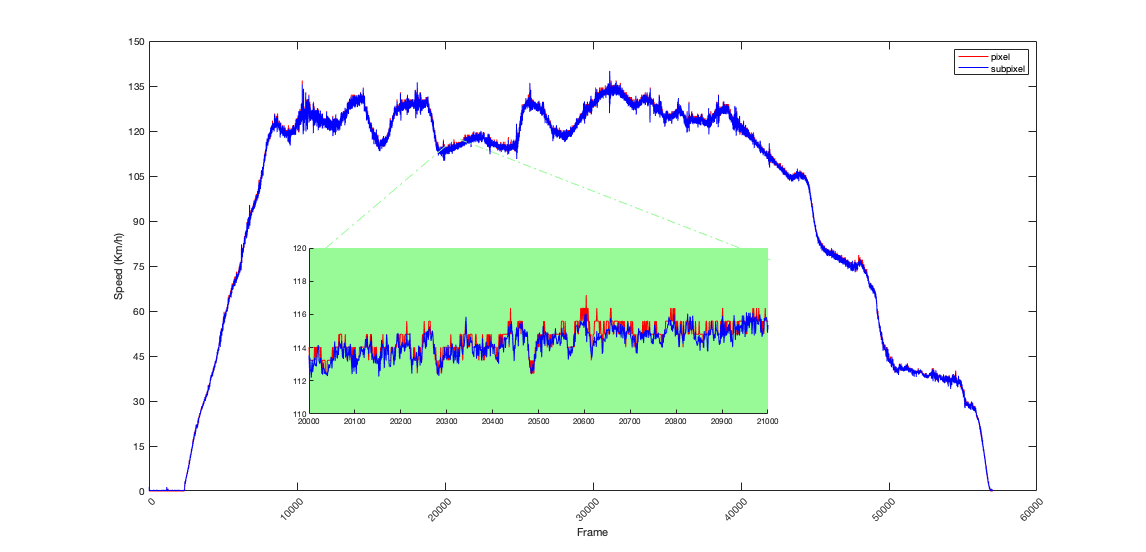}
\includegraphics[width=0.3\columnwidth]{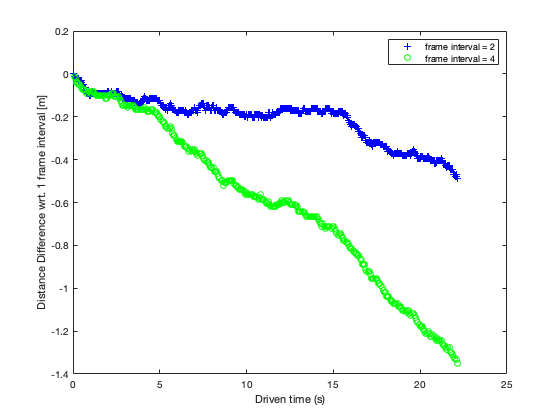}
\caption{\label{mousedist:fig}(Left) Overlay of estimated route (blue)
and GPS estimate(red);(middle)Velocity plot from the Optical Correlation
(``Train Mouse'') Module. The steps due to the pixel quantization show
the necessity for sub-pixel processing (blue) compared to pixel-accurate
SSD only method (red);
(right) Reduction of the drift accumulation through longer keyframe
sequences in optical correlation (``Train Mouse'').}
\end{figure}

The estimated profiles tracked over one of our test runs are depicted in
Fig.~\ref{mousedist:fig}. The tracked velocity was confirmed with GNSS
measurements in areas, where the GNSS reception was available. We show
the comparison in Fig.~\ref{mousedist:fig}middle.

The extension of the number of frames, in which the same template is
tracked results in significant reduction of drifts
(Fig.~\ref{mousedist:fig})right.
\begin{figure}[ht]
\centering
\includegraphics[width=\columnwidth]{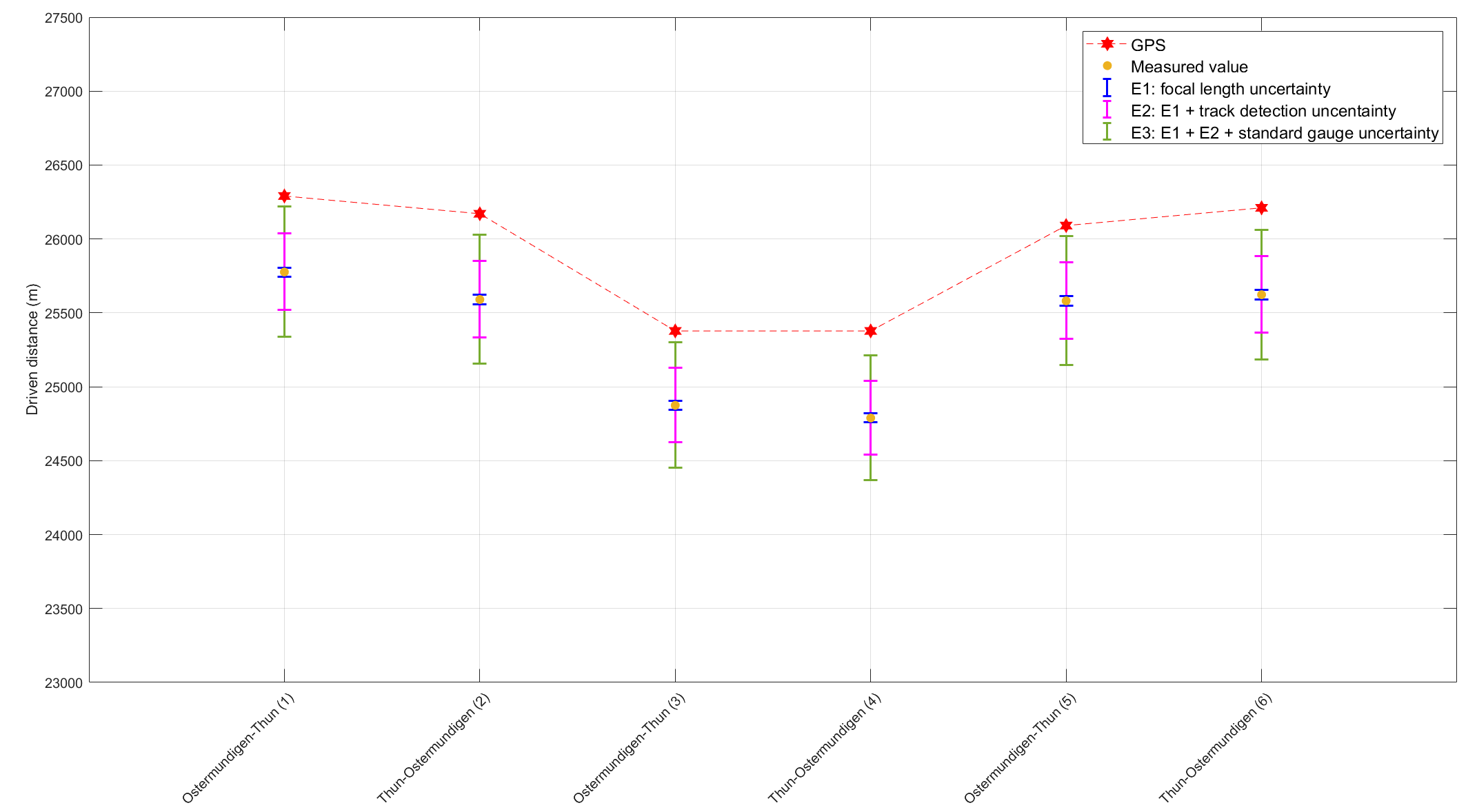}
\caption{\label{drift2:fig} 
Agreement between the GPS measurement and distance estimated with
``Train Mouse''.} 
\end{figure}

The additional drift in accumulated over a distance of 855m was 0.84m
for a system with 1/4 less reference frame switches. In 22 seconds the
system made 30*22 switches for the (green) case of continuous reference updates
and 15*22 switches for the (blue) case of a keyframe sequence length of
4. The decrease in the measured distance is -1.31m for the blue and
0.49m in the green case. We see that the blue curve traveled a shorter
distance than the continuously switching case which corresponds to the
zero line in Fig.~\ref{mousedist:fig}right.

Fig.~\ref{drift2:fig} shows a good estimate of the traveled distance
compared to the GPS measurement based entirely on the ``Train Mouse''
without compensation with April-Tags. We see how different parameters, like
focal length, height above the ground, and gauge distance influence the
parameters. We currently work on on-line re-calibration of this error. 
\subsection{Performance of the Motion Estimator}

The system was run on a Quadcore Pentium i5\@ 3.1GHz with an NVIDIA
GTX1080 for low-level image processing. The system was able to do
per-frame calculations in the range 07-11ms/frame, which allows online
processing of the 60Hz image streams from the camera. 

\begin{figure}[ht]
\centering
\includegraphics[height=2.6cm]{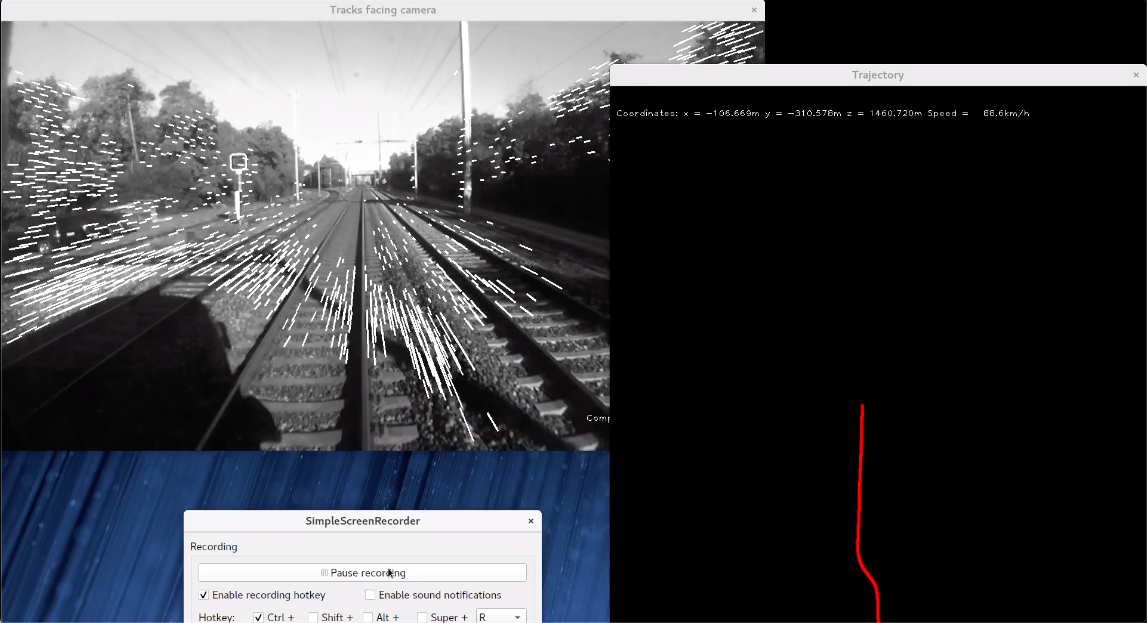}
\includegraphics[height=2.6cm]{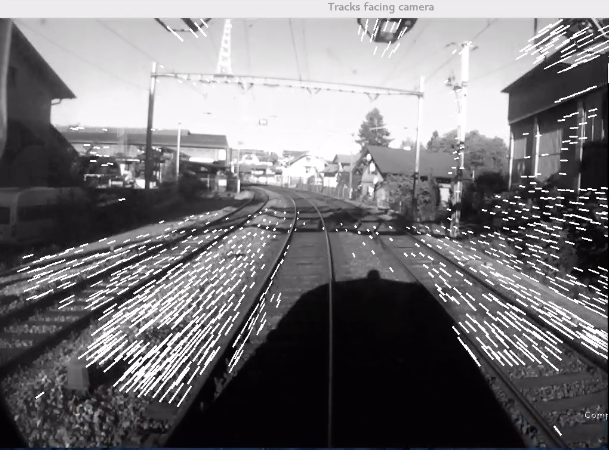} 
\caption{\label{visodo:fig} Real-time calculation of filtered flow for
higher velocities (left) and curve motion (right).}
\end{figure}

Fig.~\ref{visodo:fig} shows the system processing along the route for
the case of an open space and a curve motion. The map plotted in red
next to the visualization window corresponds to the route form from the
local region. The accuracy of this part of the processing was already
successfully verified in~\cite{darius_loc_chapter}. Our current test was
to see, how many optical flow vectors can explain the current motion of
the train but passing the epipole not further than 0.5 pixels. The while
line segments in Fig.~\ref{visodo:fig} shows a very large number of such
segments with a large spread over the image. This results in a very
small drift in motion orientation~\cite{elmarIros}.

We see in Fig.~\ref{visodo:fig} that the epipole prediction from the
computation in the Optical Correlation (``Train Mouse'') module can
successfully be used to filter correct correspondences that capture the
ego motion of the train with the point of expansion in the intersection
point of the ego-velocity vector with the image plane (epipole).

In comparison to current system like the ones in ~\cite{siegwart}, we can keep up with speeds greater than 140 km/h. There is nearly no influence from other objects, since we only rely on a small track area in front of the rail vehicle. Using NIR cameras in our approach weakens the effect of shadows and sunlight. With the “train mouse” we derive the gain for the z-axis directly from the known track gauge width, avoiding gain drifts as in ~\cite{siegwart}.

\subsection{Fusion of Navigation Data from Global Landmarks}
Using sensor fusion techniques (e.g. Kalman filtering based on a train model) the relative positions can be combined with the world coordinates thus effectively compensating for drift and filter off outliers and noise. Using an IMU together with the train model will provide even more robust results. In addition an accurate and trusted topological map of the tracks can be used to further improve accuracy and to determine the integrity of the position information.    

\begin{figure}[ht]
\centering
\includegraphics[width=8cm]{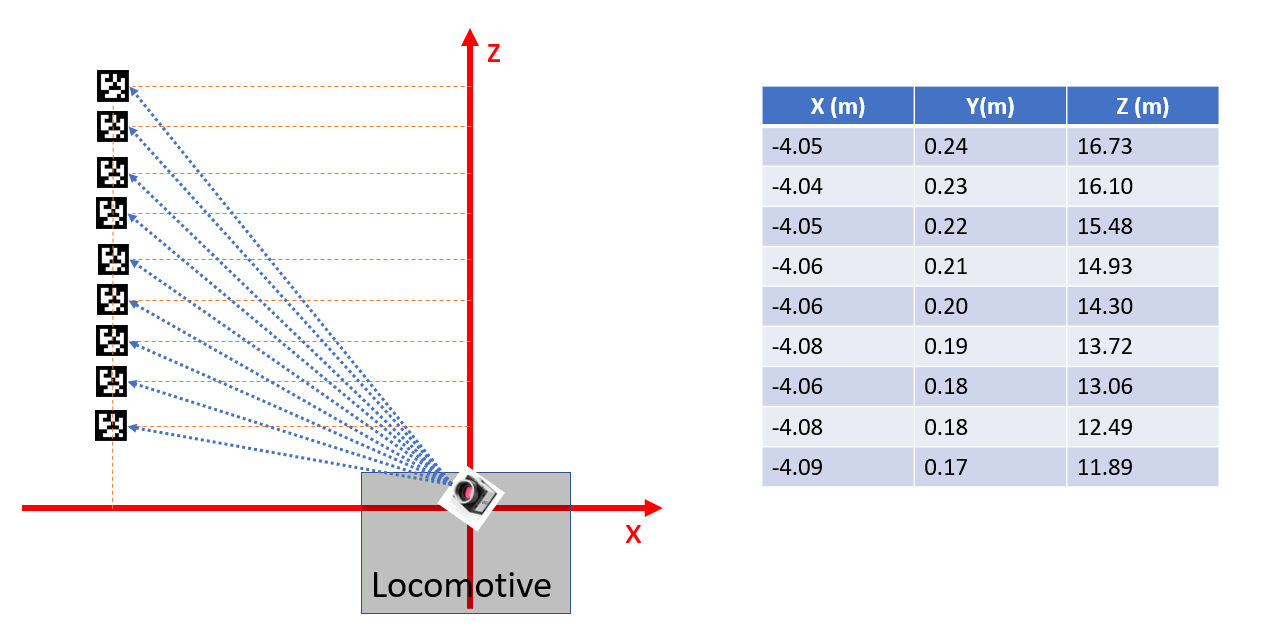}
\caption{\label{tag_results:fig} Results of pose estimation for an April tag.}
\end{figure}

In Figure~\ref{tag_results:fig} the pose estimation results for a given April tag when passing by with the measurement train are shown. It can be seen that the perpendicular distance from the pole to the track (X) is nearly constant as well as the height (Y) while passing an April tag mounted on a pole of the catenary. The table shows the values for 9 consecutive image frames while the train covers a distance of approx. 5 m.  

\section{Conclusions and Future Work}
\label{conclusion:sec}


We presented a system that represents an approach to deal with specific
requirements of unstructured dynamic railroad scenarios. The system
shown in Fig.~\ref{system:fig} has a modular structure with modules
that provide the dynamic motion updates with varying update rates and
drift properties. In our current application, the train dynamics is slow
enough due to the stiff suspension of the trains to observe the dynamics
with the 60Hz update rate of our monocular camera system. We plan to
extend it to more agile train suspensions that will require a faster
dynamic update and an addition of an  IMU unit shown in the
Fig~\ref{system:fig}. 

Our main contribution is three-fold:

\paragraph{Low-Level Matching under Strong Self-Smilarity} - we adapted
the low-level vision unit to cope with the ambiguous world of the
railroad environment with very strong self-similarity between the local
objects (stones, screws, etc.).  We extended the local descriptor to the
entire track area under the planarity assumption for the track bed. This
makes the matching system more robust. The vanishing points from the
motion estimate allow also a robust filtering of correct landmarks for
SfM module without any random selections as it is the case for RANSAC
systems. This is an essential pre-requisite to be able to make the
system verifiable for the SIL 4.0 requirements.  

\paragraph{Calculation
of Error Covariance} - our system calculates not only the current pose
change but also the confidence of the result as a covariance matrix.
This allows on one hand a better monitoring of the QoS (Quality of
Service) but at the same time it improves essentially the convergence
properties of the fusion network in the Navigation Unit
(Fig.~\ref{system:fig}). The processing allows also weighing of the used
optical flow vectors depending on their reliability (length in the
image). This is our next step to improve the accuracy of the SfM module.
\paragraph{Key-frame Processing} - instead of a bundle-adjustment step
common for most of the SLAM approaches, we apply the key-frame
processing idea that reduces the number of reference changes during
operation of the unit. The specific problem of railroad environments is
the strong occlusions of distant features which requires to focus on the
track-bed itself as navigation area. We currently compensate for drifts
with artificial landmarks, e.g., April tags along the way, but we plan
to use a system that will try to re-identify distant objects (once they
come into view after a train pass again) that will also allow to
compensate for the drift more efficiently.

With the above shown “optical navigation” the following properties were achieved for the visual sensors:

1.	skid-free odometer (visual odometry)
2.	“visual” balise (detecting April tags) 
3.  incremental motion
4.	track selectivity
5.	global pose reference in six dimensions (combined with a map) (visual localization)
6.  real-time capability of the image processing @ 60 Hz

The “optical navigation” presented here has the advantage that it is deterministic and  does not require machine learning algorithms (for example neural networks or deep learning). No SIL4 application based on machine learning has yet been approved by a relevant certification body. 
This “optical navigation” opens up the opportunity to provide evidence of a safe and secure image processing for the localization. Since the image processing can provide incremental motion as well as global pose reference it is a potential sensor for a safe sensor-fusion, but to guarantee diversity it must be combined with other sensors (for example IMU). 

Interoperability within the European railways and setting international standardization has to be achieved, after the sensor-fusion is proven and the first approval through a relevant certification body was accomplished . 

The data acquisition took place in a vision friendly environment with good weather conditions. Therefore dealing with worse weather conditions or other environments is beyond the scope of this paper and it will be focused on in the next steps.

\bibliographystyle{plain}
\bibliography{lit}
\end{document}